\documentclass[letterpaper, 10pt, conference]{ieeeconf}

\IEEEoverridecommandlockouts
\usepackage{graphicx}
\usepackage{booktabs}
\usepackage{amsmath}
\usepackage{amssymb}
\usepackage{algorithm}
\usepackage{algorithmic}
\usepackage{cite}
\usepackage{url}
\usepackage[hidelinks,bookmarks=false]{hyperref}
\usepackage{multirow}

\hypersetup{
  pdftitle={Paired-CSLiDAR: Height-Stratified Registration for Cross-Source Aerial-Ground LiDAR Pose Refinement},
  pdfauthor={Montana Hoover, Jing Liang, Tianrui Guan, Dinesh Manocha}
}

\title{\LARGE \bf
Paired-CSLiDAR: Height-Stratified Registration for\\Cross-Source Aerial--Ground LiDAR Pose Refinement}

\author{%
Montana Hoover$^{1}$, Jing Liang$^{2}$, Tianrui Guan$^{1}$, Dinesh Manocha$^{1}$%
\thanks{$^{1}$University of Maryland, $^{2}$Stanford University. \texttt{mjhoover@umd.edu}}%
}

\begin{document}

\maketitle
\thispagestyle{empty}
\pagestyle{empty}

\begin{abstract}
We introduce \emph{Paired-CSLiDAR} (CSLiDAR), a cross-source aerial--ground LiDAR benchmark for single-scan pose refinement---refining a ground-scan pose within a 50\,m-radius aerial crop---with \textbf{12,683} ground--aerial pairs across \textbf{6} evaluation sites and per-scan reference 6-DoF alignments for sub-meter root-mean-square error (RMSE) evaluation.
Because aerial scans capture rooftops and canopy while ground scans capture facades and under-canopy, the two modalities share only a fraction of their geometry---primarily the terrain surface---causing standard registration methods (iterative closest point (ICP) variants and learned correspondence models) to converge to metrically incorrect local minima.
To address this, we propose a training-free, geometry-only refinement pipeline, \textbf{Residual-Guided Stratified Registration (RGSR)}, that exploits the shared ground plane as the most reliably overlapping structure: height-stratified ICP generates diverse candidate poses from varying ground-plane subsets and reversed registration directions, and confidence-gated accept-if-better selection ensures no RMSE regressions.
RGSR achieves \textbf{86.0\%} S@0.75\,m and \textbf{99.8\%} S@1.0\,m on the primary benchmark (9,012 scans), outperforming both the confidence-gated cascade (83.7\%) and a learned baseline (GeoTransformer, 76.3\%).
We validate RMSE-based pose selection via independent survey control and trajectory consistency; a separate Fourier--Mellin BEV analysis reveals that added proposals can lower RMSE while increasing actual pose error under extreme partial overlap, highlighting the need for non-circular validation.
The pipeline requires no training and provides a diagnostic baseline for the benchmark; the dataset (CC~BY~4.0) and code (MIT) are being prepared for public release.
\end{abstract}

\section{INTRODUCTION}

Reliable localization underpins autonomous mobile robots and is commonly achieved by LiDAR SLAM (simultaneous localization and mapping) or scan-to-map registration within a prior map~\cite{cadena2016slam,pomerleau2015review,loam,liosam,icp,ndt}.
Publicly available airborne LiDAR data~\cite{usgs3dep} can serve as prior maps for ground robots, but the aerial--ground viewpoint gap is recognized in robot localization with overhead 3D priors~\cite{vandapel2006ugv,madhavan2005temporal} and in airborne--mobile/terrestrial laser scanning (ALS--MLS/TLS) alignment~\cite{teo2014alsmls,yang2015alstls}.

\begin{figure}[t]
  \centering
  \includegraphics[width=\columnwidth]{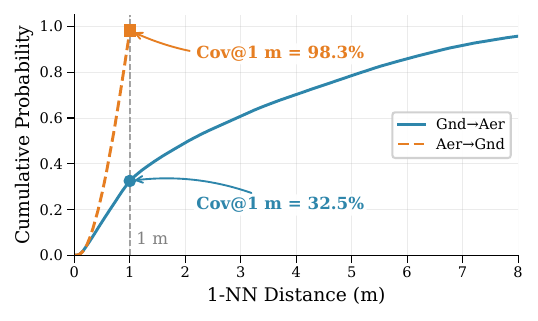}
  \caption{%
    \textbf{Coverage asymmetry} (UMD subset, 190 scans pooled).
    Aerial LiDAR captures rooftops and top-canopy while ground LiDAR captures facades and under-canopy, so many ground points have no aerial counterpart.
    Cumulative distribution (CDF) of nearest-neighbor (1-NN) distances between paired clouds under $T_{\mathrm{ref}}$; Cov@1\,m (fraction of \emph{source} points with a neighbor ${\leq}$1\,m) is marked on each curve.
    The metric is directional: ground${\to}$aerial Cov@1\,m${=}$32.5\% (per-site means 20--68\%; Table~\ref{tab:dataset}), while aerial${\to}$ground reaches 98.3\%.
    Paired-CSLiDAR reports this directionality, and RGSR exploits it by reversing ICP direction (aerial as source, then invert and refine; Sec.~\ref{sec:rgsr}) to reach different local minima.
  }
  \label{fig:coverage}
\end{figure}

The core difficulty in cross-source aerial--ground registration is \emph{extreme, directional partial overlap} (Fig.~\ref{fig:coverage}): aerial maps capture rooftops and top-canopy while ground scans capture facades and under-canopy, and only the terrain is reliably shared between the aerial and ground. As a result, many ground points lack an aerial counterpart.
We quantify this via Cov@1\,m, the fraction of \emph{source} points whose nearest neighbor in the \emph{paired} scan lies within 1\,m (a distance comparable to the aerial point spacing at 2--8\,pts/m$^2$) under the per-scan reference alignment $T_{\mathrm{ref}}$.
This overlap is highly asymmetric: when ground points are queried against the aerial map, Cov@1\,m can be as low as 20\% because facades and under-canopy have no aerial counterpart; conversely, nearly all aerial points (98.3\%; Fig.~\ref{fig:coverage}) find a ground neighbor, because the terrain surface is observed from both viewpoints (Table~\ref{tab:dataset}).
As a nearest-neighbor local optimizer, ICP~\cite{icp} can converge to incorrect local minima when many source points match to non-overlapping regions~\cite{pomerleau2015review,tricp}; our refinement pipeline (RGSR; Sec.~\ref{sec:rgsr}) addresses this by generating diverse hypotheses that exploit the high aerial-to-ground overlap direction and the shared ground plane, selecting the best pose via accept-if-better RMSE guards.
Two learned baselines (GeoTransformer~\cite{geotransformer}, BUFFER-X~\cite{bufferx}) also degrade in these low-overlap settings (Sec.~\ref{sec:experiments}); RGSR outperforms the best learned method by 9.7\,pp at S@0.75\,m (Table~\ref{tab:rgsr}), indicating that the coverage asymmetry---not only cross-source representation shift---drives registration failures.

\emph{Task definition:}
We investigate the problem of \emph{single-scan metric pose refinement} (sub-meter inlier RMSE; not global localization): given a ground scan $S$, its paired 50\,m-radius aerial crop $A$ (chosen as a practical submap size; e.g., 100${\times}$100\,m aerial patches in CrossLoc3D~\cite{crossloc3d}, similar scale to other cross-view submaps~\cite{hotformerloc}), and a coarse prior $T_{\mathrm{init}}$ (e.g., GNSS+IMU or aerial--ground retrieval~\cite{crossloc3d,hotformerloc}) that selects the \emph{correct} crop, we estimate a refined 6-DoF (degrees of freedom) transform $T^*\in\mathrm{SE}(3)$ (3D rigid transform). Our formulation takes into account two protocols.
In Protocol~A, we set $T_{\mathrm{init}}{=}T_{\mathrm{ref}}$ to isolate cross-source registration difficulty from initialization noise; in Protocol~B, we perturb $(x,y,\mathrm{yaw})$ by $\pm$5\,m and $\pm$15$^\circ$ (holding $z$, roll, pitch fixed) to benchmark convergence to the correct alignment within the correct crop.
This reflects gravity-aligned priors: roll/pitch are constrained; planar/yaw can still fail under low overlap.
We target $\tau{=}0.75$\,m inlier RMSE as the primary sub-meter threshold because it serves as our cascade gate and correlates with independent survey-based pose error, as shown in Sec.~\ref{sec:analysis}.

\emph{Main results:}
We present a novel benchmark and a refinement pipeline.
Our formulation exploits the \emph{ground-plane structure} as the most reliably shared geometry across viewpoints, providing well-conditioned constraints on $z$, roll, and pitch.
Building on standard coarse-to-fine (CTF) ICP, we generate multiple candidate poses from diverse initializations (varying height-percentile subsets and source/target direction) and select conservatively via \emph{accept-if-better} (accept only if post-ICP inlier RMSE decreases).
We report Success@$\tau$ (S@$\tau$): the fraction of scans with inlier RMSE${<}\tau$, and validate RMSE-based selection with survey control and trajectory consistency (Sec.~\ref{sec:analysis}).

\textbf{Contributions:} We make three contributions:
\begin{enumerate}
  \item \emph{Benchmark (Paired-CSLiDAR).} We introduce a public cross-source aerial--ground LiDAR benchmark with per-scan reference $\mathrm{SE}(3)$ alignments for sub-meter RMSE evaluation---\textbf{12,683} ground--aerial pairs and 50\,m aerial crops across \textbf{6} evaluation sites for single-scan pose refinement (Sec.~\ref{sec:dataset}), enabling reproducible, overlap-stratified evaluation with per-scan reference alignments.
  \item \emph{Method (RGSR geometry-only pipeline).} We propose a training-free, non-regressive refinement pipeline that reaches diverse ICP local minima via height stratification and reverse-direction hypotheses, with accept-if-better RMSE selection (Sec.~\ref{sec:two_stage}--\ref{sec:rgsr}). \textbf{RGSR} reaches \textbf{86.0\%} S@0.75\,m on \textbf{9,012} Protocol~B scans, outperforming GeoTransformer+CTF by 9.7\,pp (Table~\ref{tab:rgsr}), showing that directional overlap (not only representation shift) drives failures.
  \item \emph{Evaluation (non-circular validation).} We validate post-ICP RMSE as a selection signal via survey control ($\rho{=}0.49$ at CTF stage, 200 scans) and trajectory consistency, and we identify a metric disconnect---added hypotheses can lower RMSE while increasing pose error---as a benchmark-level finding (Fourier--Mellin BEV; Sec.~\ref{sec:analysis}).
\end{enumerate}

\section{RELATED WORK}

\textbf{Registration under partial overlap.}
Classical methods include ICP (iterative closest point)~\cite{icp}, NDT (normal distributions transform)~\cite{ndt}, and feature-based initialization (FPFH (fast point feature histograms) + RANSAC (random sample consensus)~\cite{rusu2009fast,ransac}) refined by ICP.
Trimmed ICP~\cite{tricp} addresses partial overlap by discarding worst correspondences; GICP~\cite{gicp} improves local convergence via covariance modeling but remains vulnerable when overlap is extremely low or highly asymmetric.
Overlap-estimation methods~\cite{eoe,stetchschulte2019} target robustness under limited shared geometry; our aerial--ground setting poses additional challenges because the unmatched fraction is highly directional (Sec.~\ref{sec:dataset}).
Standard ICP formulations are asymmetric (source${\to}$target); symmetric objectives have been proposed~\cite{rusinkiewicz2019symmetric} but do not address directional overlap. Under such overlap, swapping roles changes which points dominate the objective, a property we exploit (Sec.~\ref{sec:rgsr}).
Global solvers include Super4PCS~\cite{super4pcs}, FGR~\cite{fgr}, and TEASER++~\cite{teaser}; correlative scan matching~\cite{olson2009} generates bird's-eye view (BEV) hypotheses, which our exploratory spectral extension (+FM; Sec.~\ref{sec:rgsr}) adapts via Fourier--Mellin phase correlation~\cite{reddy1996fft}.
In contrast to overlap-robust ICP variants that assume roughly symmetric overlap, RGSR explicitly uses direction swapping as a hypothesis generator.

\textbf{Learning-based and cross-source registration.}
Deep correspondence models~\cite{geotransformer} are typically trained and evaluated on same-source benchmarks (e.g., 3DMatch~\cite{threedmatch}).
Cross-source methods such as BUFFER-X~\cite{bufferx}, CrossPCR~\cite{crosspcr}, and SPEAL~\cite{speal} address density and sensor variation, but aerial--ground registration additionally introduces large viewpoint/occlusion differences and coverage asymmetry; the two learned baselines we evaluate (GeoTransformer trained in-domain and BUFFER-X zero-shot) do not outperform CTF and collapse on the lowest-coverage sites (Tables~\ref{tab:cascade},~\ref{tab:rgsr}).
We include learned baselines diagnostically, to isolate whether failures stem from representation shift or from directional overlap asymmetry that also affects classical methods.

\textbf{Aerial--ground localization and benchmarks.}
Registering ground range data to overhead 3D priors for unmanned ground vehicle (UGV) localization has been explored~\cite{vandapel2006ugv,madhavan2005temporal}.
CrossLoc3D~\cite{crossloc3d} and HOTFormerLoc~\cite{hotformerloc} provide aerial--ground datasets but target \emph{retrieval} (coarse spatial tolerances); recent work incorporates coarse-to-fine registration for metric localization~\cite{hotflocpp} but without pre-paired crops with per-scan reference transforms or sub-meter evaluation.
To our knowledge, no public benchmark defines a single-scan LiDAR-to-aerial-LiDAR metric refinement task with (i)~pre-paired crops, (ii)~per-scan reference alignments, and (iii)~sub-meter RMSE-threshold evaluation.
USGS 3D Elevation Program (3DEP)~\cite{usgs3dep} covers much of the continental US; our prior GND dataset~\cite{gnd} collected ground-robot LiDAR across university campuses but without paired airborne crops or per-scan metric references.
Existing single-source benchmarks (KITTI~\cite{kitti}, NCLT~\cite{nclt}, MulRan~\cite{mulran}), ALS--MLS/TLS registration studies~\cite{teo2014alsmls,yang2015alstls}, and cooperative datasets~\cite{agcdrive,graco} lack the refinement task considered here.
Paired-CSLiDAR complements retrieval-focused datasets by protocolizing scan-to-crop pairing with per-scan reference transforms for consistent sub-meter refinement evaluation.

\section{DATASET: PAIRED-CSLIDAR}
\label{sec:dataset}

\subsection{Data Collection}

\begin{figure*}[t]
  \centering
  \includegraphics[width=\textwidth]{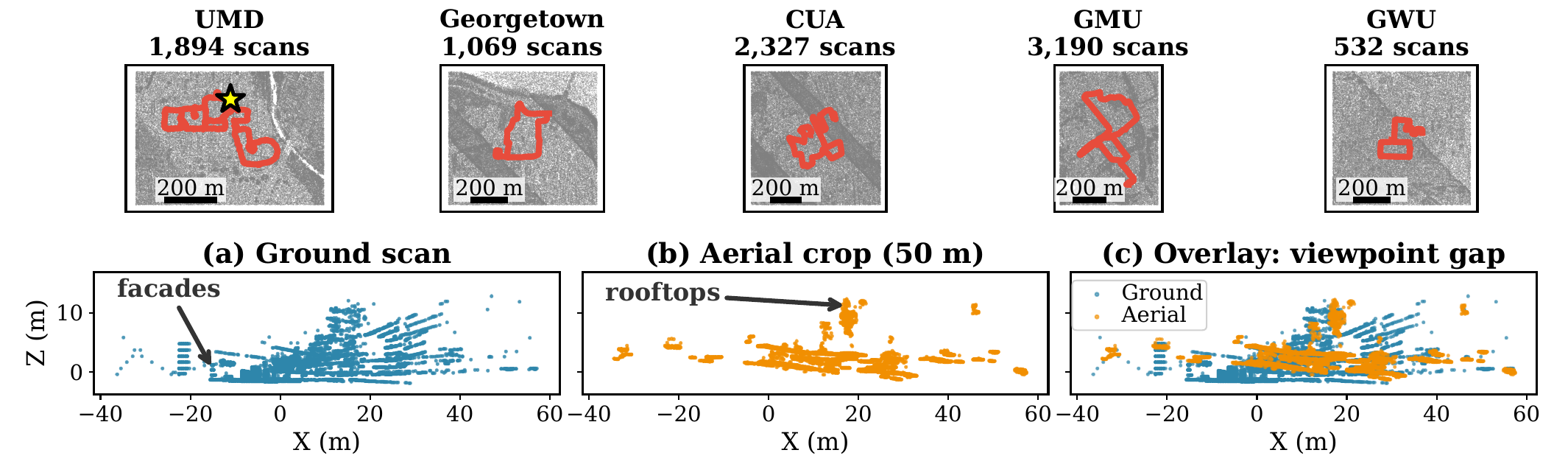}
  \caption{%
    \textbf{Dataset overview.}
    \emph{Top:} Bird's-eye view of UMD, Georgetown, CUA, GMU, and GWU with ground trajectories (red) overlaid on airborne LiDAR maps (gray); scale bars show 200\,m. UMD includes two routes (IdeaFactory / Iribe). Scan counts correspond to the Protocol~B primary benchmark (9{,}012 scans); full dataset statistics (12{,}683) are in Table~\ref{tab:dataset}.
    \emph{Bottom:} One paired sample from UMD (star in top panel; side view): ground scan~(a), 50\,m-radius aerial crop~(b), and their overlay~(c) under $T_{\mathrm{ref}}$, illustrating the viewpoint gap and missing facade/sub-canopy structure in aerial maps.
  }
  \label{fig:dataset}
\end{figure*}

We collected ground LiDAR scans at 5 US universities using mobile unmanned ground vehicles (UGVs) with spinning LiDAR (Velodyne VLP-16 at 4 sites; Ouster OS1-64 at UMD).
SC-LIO-SAM~\cite{scliosam} provided deskewed 360$^\circ$ keyframe sweeps and SLAM trajectories; we retain every keyframe (${\approx}$1.2\,m spacing) and intend to release each as a native-density XYZ point set in its scan frame.
We obtained aerial LiDAR from publicly available sources: MD iMAP~\cite{mdimap} (Maryland), USGS 3DEP~\cite{usgs3dep} (District of Columbia), and Virginia Geographic Information Network (VGIN)~\cite{vavgin} (Virginia), at nominal acquisition densities of 2--8\,pts/m$^2$.
Table~\ref{tab:dataset} summarizes the dataset statistics.
GMU comprises five trajectories and Georgetown two (aggregated in Table~\ref{tab:dataset}; similar within-campus coverage); UMD contributes two disjoint routes with contrasting coverage (20\% vs.\ 46\%), yielding 6 evaluation sites (Fig.~\ref{fig:dataset}).
The planned public release will include the ground scans plus per-scan pairing, odometry, and reference-pose metadata under CC~BY~4.0 and the code under MIT; aerial crops are reproduced from the original public tiles (exact product/tile IDs recorded) via provided download+crop scripts (provider licenses apply).

\begin{table}[t]
\centering
\caption{\textbf{Dataset statistics.} 12,683 scan pairs across 6 evaluation sites. Cov@1$\,\mathrm{m}$: mean (over scans) fraction of ground-scan points whose 3D nearest neighbor in the aerial crop, under $T_{\mathrm{ref}}$, lies within $1\,\mathrm{m}$. UMD is split into IdeaFactory and Iribe because the two routes have markedly different coverage (20\% vs.\ 46\%). Protocol~B per-site counts (after exclusions) are in Table~\ref{tab:rgsr}. The wide 20--68\% coverage range enables overlap-stratified evaluation of refinement difficulty.}
\label{tab:dataset}
\small
\setlength{\tabcolsep}{2.5pt}
\begin{tabular}{@{}l r r c r@{}}
\toprule
Site & Scans & Traj.\ len. & Aerial src.\ & Cov@1\,\textnormal{m} \\
\midrule
GMU$^\ast$ & 5,361 & 6.4\,km & VA VGIN & 65\% \\
Georgetown$^\ast$ & 2,569 & 3.0\,km & USGS 3DEP & 61\% \\
CUA & 2,327 & 2.2\,km & USGS 3DEP & 68\% \\
UMD (IdeaF.) & 1,031 & 1.5\,km & MD iMAP & 20\% \\
UMD (Iribe) & 863 & 1.1\,km & MD iMAP & 46\% \\
GWU & 532 & 0.8\,km & USGS 3DEP & 66\% \\
\midrule
\textbf{Total} & \textbf{12,683} & \textbf{15.0\,km} & --- & \textbf{60\%}\\
\bottomrule
\multicolumn{5}{l}{\scriptsize $^\ast$Includes Protocol~B exclusions (Georgetown: 1{,}500; GMU: 2{,}171).}
\end{tabular}
\end{table}

\subsection{Pairing and Reference Poses}

For each ground scan $S_i$ (full 360$^\circ$, native-density XYZ), the paired aerial submap $A_i$ contains all aerial points within 50\,m in $(x,y)$ of the scan's reference translation (all provider returns; noise class removed where present).
We provide $T_{\mathrm{ref},i}\in\mathrm{SE}(3)$ mapping each scan's native frame into a site-local aerial frame (meters, $z$-up; provider coordinate reference system (CRS) translated to a local origin).
Ground${\to}$aerial Cov@1\,m ranges from 20\% to 68\% (overall mean 60\%); the uncovered fraction observes facades and sub-canopy geometry invisible to the aerial sensor.

\textbf{Reference alignments.}
We aligned each SLAM trajectory map to the aerial point cloud via manual correspondence selection (stable built structures) followed by ICP refinement, producing one rigid SE(3) per trajectory; per-scan $T_{\mathrm{ref},i}$ are obtained by composing this trajectory-level transform with the SLAM pose.
Each alignment was verified by visual inspection of building-scale structures; the planned public release will include correspondence points and trajectory-level transforms to enable exact reproduction.
Cross-source NN residuals exhibit a nonzero floor because many ground points have no aerial counterpart (per-site median ground${\to}$aerial NN distance under $T_{\mathrm{ref}}$: 0.3--3.6\,m, reflecting coverage variation); we therefore use ``reference'' rather than ``ground truth.''
Reference poses define aerial crop pairing and $T_{\mathrm{init}}$ ($T_{\mathrm{ref}}$ in Protocol~A; jittered in Protocol~B).
Inlier RMSE $e(T)$ is used as a selection score; pose accuracy is validated via survey control (200 UMD scans; $\rho{=}0.49$ at CTF stage, $p$-value${<}2{\times}10^{-13}$) and trajectory consistency (Sec.~\ref{sec:analysis}).

\textbf{Quality filtering.}
Protocol~B excludes three trajectories from Protocol~A's 12{,}683 pairs:
one Georgetown route (1{,}500 scans) with SLAM drift artifacts that render aerial crops unreliable under jitter, and two overlapping GMU routes (2{,}171 scans) whose trajectories spatially overlap with the three retained GMU trajectories.
\emph{Filtering uses only trajectory-level criteria (drift integrity, spatial independence), never RMSE or any registration result.}

\textbf{Primary benchmark.}
Protocol~B ($\pm$5\,m, $\pm$15$^\circ$ jitter) on 8 non-overlapping trajectories (of 11 total) yields \textbf{9{,}012 scans across 6 sites} (Table~\ref{tab:rgsr}).
Protocol~A (Table~\ref{tab:cascade}) uses all 12{,}683 pairs.

\section{METHOD: HEIGHT-STRATIFIED REGISTRATION}
\label{sec:method}

\begin{figure}[t]
  \centering
  \includegraphics[width=\columnwidth]{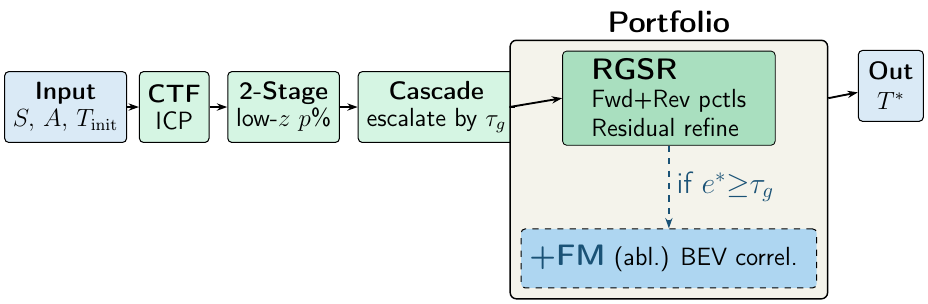}
  \caption{%
    \textbf{Registration pipeline.} \emph{Cascade} (CTF $\to$ Two-Stage $\to$ RANSAC+CTF as needed) escalates by RMSE threshold~$\tau_g$. \textbf{RGSR} extends the cascade with 8 Two-Stage hypotheses (4~percentiles $\times$ \{fwd, rev\}) plus residual refinement. +FM (exploratory; Sec.~\ref{sec:rgsr}) optionally adds spectral BEV proposals when RMSE~${\geq}\tau_g$. All transitions use \mbox{accept-if-better} selection, yielding stronger low-coverage performance without RMSE regressions.
  }
  \label{fig:pipeline}
\end{figure}

Our approach exploits a structural property of aerial--ground LiDAR: the ground plane is mutually observed and provides well-conditioned constraints on vertical translation, roll, and pitch (Fig.~\ref{fig:pipeline}).
RGSR is designed as a \emph{diagnostic baseline} that decomposes the cross-source registration problem into interpretable, independently measurable stages (Table~\ref{tab:rgsr}), enabling future methods to target specific failure modes (e.g., canopy-dominated sites, low-overlap settings).
All candidate poses are scored by \emph{inlier RMSE} $e(T)$, always computed in the forward direction ($S$ as source, $A$ as target):
let $\mathcal{I}(T){=}\{i : \|Ts_i{-}\mathrm{NN}_A(Ts_i)\|{\leq}r_{\mathrm{eval}}\}$; then $e(T){=}\sqrt{|\mathcal{I}|^{-1}\sum_{i\in\mathcal{I}}\|Ts_i{-}\mathrm{NN}_A(Ts_i)\|^2}$ if $|\mathcal{I}|{\geq}50$, else $\infty$ ($r_{\mathrm{eval}}{=}2.0$\,m).
This score is used for cascade escalation and accept-if-better selection; reverse-direction hypotheses are inverted and re-scored in this same forward direction.

\subsection{Two-Stage Coarse-to-Fine ICP}
\label{sec:two_stage}

Standard coarse-to-fine (CTF) ICP applies decreasing distance thresholds $[5, 3, 2, 1.5, 1]$\,m, but facade and vegetation residuals bias ICP under coverage asymmetry.
\textbf{Key insight:} Ground-plane points are approximately planar and mutually observed from both viewpoints.
Two-Stage CTF bootstraps from the lowest $p$-th height percentile, then refines on the full cloud.
Heights $z_i{=}(T_{\mathrm{init}} s_i)_z$ are computed once in the aerial map frame (gravity-aligned $z$) and the subset $S_p{=}\{s_i : z_i \leq \mathrm{quantile}_p\}$ is held fixed during the coarse stage; for reverse-direction hypotheses, percentiles use the aerial cloud's native $z$.
\begin{enumerate}
  \item \emph{Coarse stage:} unconstrained SE(3) ICP on the lowest-$p$ percentile \textbf{of the source scan} (aerial target unrestricted) with thresholds $[5, 3, 2]$\,m, biasing updates toward $(z, \mathrm{roll}, \mathrm{pitch})$ via the near-planar subset.
  \item \emph{Fine stage:} ICP on all source points with thresholds $[2, 1.5, 1]$\,m, resolving $(x, y, \mathrm{yaw})$ with full-scene geometry.
\end{enumerate}
The percentile $p$ trades constraint quality against quantity; $p{=}30$ balances across site types.

Gains are largest on the lowest-overlap sites, where ground-plane bootstrapping stabilizes vertical/tilt before optimizing $(x, y, \text{yaw})$ (per-site results in Table~\ref{tab:rgsr}).

\subsection{Confidence-Gated Cascade}
\label{sec:cascade}

The cascade escalates to more expensive methods only when RMSE exceeds $\tau_g{=}0.75$\,m, a gate justified by the correlation between inlier RMSE and survey-control translation error (Spearman $\rho{=}0.49$, $p$-value${<}2{\times}10^{-13}$; Sec.~\ref{sec:analysis}):

\begin{enumerate}
  \item Run CTF. If $e(T) < \tau_g$, accept and stop.
  \item Otherwise, run Two-Stage CTF ($p{=}30$).
  \item If $e(T)$ still $\geq \tau_g$, run FPFH-RANSAC~\cite{rusu2009fast,ransac} + CTF refinement.
\end{enumerate}

Each transition is guarded by \emph{accept-if-better}: zero RMSE regressions by construction.
The cascade achieves \textbf{99.0\%} S@1.0\,m across all 12,683 scans at $\approx$1.2$\times$ compute.

\subsection{Residual-Guided Stratified Registration (RGSR)}
\label{sec:rgsr}

The cascade uses a single ground-plane percentile ($p{=}30$).
Per-scan analysis reveals the optimal percentile varies across environments.
RGSR is invoked on every scan, initializing from the cascade output; Phase~1 exits early once $e^*{<}\tau_g$:

\textbf{Phase 1: Multi-percentile ICP hypotheses (\mbox{forward+reverse}).}
For each $p \in \{15, 30, 45, 60\}$, run Two-Stage in both directions---forward ($S$ source) and reverse ($A$ source, inverted)---seeded from $T_{\mathrm{init}}$ (forward) and $T_{\mathrm{init}}^{-1}$ (reverse), not the cascade output, to diversify local minima; refine each with standard CTF and keep the hypothesis with lowest forward-direction $e(T)$.
Phase~1 stops exploring additional percentiles once $e^*{<}\tau_g{=}0.75$\,m.

\textbf{Phase 2: Residual-guided refinement.}
For scans with $0.5 < e^* < 1.0$\,m, we refine using the height band with the best-matching geometry.
Specifically, for each forward-direction inlier ($\|T^*s_i{-}\mathrm{NN}_A(T^*s_i)\|{\leq}r_{\mathrm{eval}}$), let $r_i{=}\|T^*s_i{-}\mathrm{NN}_A(T^*s_i)\|$.
We split the inlier source points into four equal-count bins by ascending height $(T^*s_i)_z$, select the bin with the lowest median $r_i$, and run tight-radius ICP ($[0.75, 0.5]$\,m) using only that bin as source ($A$ unchanged).
The update is accepted only if $e(T)$ on the \emph{full} source cloud decreases.

\textbf{Phase 3: RANSAC fallback.}
If $e^*\geq 1.0$\,m after Phases~1--2, re-run FPFH-based RANSAC (fresh random sample) followed by CTF refinement; the cascade already includes one RANSAC attempt, but stochastic variation can reach new local minima.

All transitions use accept-if-better guards: the current best is updated only if RMSE decreases, ensuring no RMSE regressions.

\smallskip\noindent\textbf{Non-regressive hypothesis augmentation.}
Under extreme partial overlap, different initializations converge to qualitatively different local minima.
We treat registration as a portfolio: from baseline $(T_b, e_b)$, generate hypotheses designed to reach \emph{different} local minima and accept-if-better: $T^*{=}\arg\min_{T\in\{T_b\}\cup\{T_k\}} e(T)$.
Non-regression ($e(T^*){\leq}e_b$) holds by construction; the nontrivial contribution is identifying sources that reliably reach new local minima and validating that lower $e(T)$ corresponds to better poses (Sec.~\ref{sec:analysis}).
Two mechanisms provide complementary diversity: (1)~\emph{height stratification} and (2)~\emph{reverse-direction ICP} (high aerial$\to$ground coverage).

\smallskip\noindent\textbf{Reverse-direction hypothesis.}
Coverage asymmetry is directional (high aerial$\to$ground vs.\ as low as 20\% forward; Fig.~\ref{fig:coverage}).
Reverse ICP runs Two-Stage with $A$ as source to obtain $T_{A{\to}S}$; we invert it to an $S{\to}A$ hypothesis, then refine with forward ICP.
Because point-to-point ICP weights source points uniformly, swapping direction changes which points dominate the objective, reaching different local minima.

\smallskip\noindent\textbf{Spectral BEV proposals (exploratory extension).}
After RGSR (Phases~1--3), if $e^*{\geq}\tau_g{=}0.75$\,m (14.0\% of scans), we optionally generate coarse SE(2) hypotheses via Fourier--Mellin phase correlation~\cite{reddy1996fft} on multi-channel BEV occupancy grids ($200{\times}200$ at 0.5\,m/pixel; channels encode occupancy in three height bands split at the 33rd and 67th $z$-quantiles of each cloud).
Log-polar phase correlation yields the top-$K_\theta{=}3$ yaw candidates ($\pm$30$^\circ$); spatial correlation yields $K_t{=}5$ translations per yaw, giving ${\leq}$15 hypotheses lifted to SE(3) and refined by CTF under accept-if-better selection.
Survey control reveals that FM-triggered RMSE improvement does not reliably translate to pose improvement under extreme partial overlap (Sec.~\ref{sec:analysis}).
We therefore report RGSR (without FM) as the primary result and include +FM as an ablation.

\smallskip\noindent\textbf{Compact RGSR portfolio.}
Initialize $(T^*,e^*)\leftarrow(T_c,e_c)$ from the cascade.
For $p\in\{15,30,45,60\}$ and direction $\{\mathrm{fwd},\mathrm{rev}\}$, run Two-Stage then CTF and keep a hypothesis only if full-cloud RMSE decreases; stop Phase~1 early if $e^*<\tau_g$.
If $0.5<e^*<1.0$, run residual-guided band refinement and accept only if better; if $e^*\geq1.0$, run one extra RANSAC+CTF fallback and accept only if better.
All candidates are scored in forward direction using full-cloud RMSE with $r_{\mathrm{eval}}{=}2.0$\,m and the ${<}50$-inlier guard.

\section{EXPERIMENTS}
\label{sec:experiments}

\subsection{Setup}

We evaluate under two protocols:
\textbf{Protocol~A} (reference initialization): $T_{\mathrm{init}} = T_{\mathrm{ref}}$, isolating the cross-source registration difficulty from initialization noise.
\textbf{Protocol~B} (planar/heading noise, \textbf{primary benchmark}): $T_{\mathrm{init}}$ is jittered by $\pm$5\,m in $(x,y)$ and $\pm$15$^\circ$ in yaw, with $(z, \text{roll}, \text{pitch})$ held fixed---a 3-DoF (planar translation and heading) jitter model reflecting gravity-aligned IMU/SLAM attitude priors and 3DEP vertical accuracy (${\approx}$10\,cm)~\cite{usgs3dep}; this isolates the dominant planar/yaw convergence failures under limited cross-source overlap.
Translation and yaw noise are sampled independently, uniformly over $[-5, 5]$\,m and $[-15^\circ, 15^\circ]$; all results use a single fixed seed (42); robustness to jitter magnitude is validated in Sec.~\ref{sec:analysis}.
Protocol~B uses 9{,}012 scans across 8 non-overlapping trajectories (Sec.~\ref{sec:dataset}).

Each ground scan is paired with a fixed 50\,m-radius aerial crop (Sec.~\ref{sec:dataset}); Protocol~B jitters only $T_{\mathrm{init}}$ within this correct crop.
We run Open3D~\cite{open3d} point-to-point ICP (50 iterations per stage) using the \emph{stage-specific} correspondence thresholds from Sec.~\ref{sec:method} (CTF: $[5,3,2,1.5,1]$\,m; Two-Stage: coarse $[5,3,2]$\,m on the height-percentile subset, then fine $[2,1.5,1]$\,m on the full cloud).
We define the inlier error $e(T)$ as the RMSE over nearest-neighbor correspondences (transformed source $\to$ target k-d tree) within $r_{\mathrm{eval}}{=}2.0$\,m on the full source cloud; if fewer than 50 inliers exist, $e(T){=}\infty$.
Success@$\tau$ (S@$\tau$) measures the fraction of scans with $e(T){<}\tau$.
Accept-if-better selection (Sec.~\ref{sec:rgsr}) uses the same $e(T)$, so adding hypotheses cannot decrease S@$\tau$.
We report S@$\tau$ because $e(T)$ is the score used by the portfolio; survey control and trajectory consistency indicate it is a meaningful pose proxy within RGSR, but it can fail under extreme partial overlap (Sec.~\ref{sec:analysis}).
Note: coverage statistics (Fig.~\ref{fig:coverage}, Table~\ref{tab:dataset}) use $r_{\mathrm{cov}}{=}1.0$\,m to characterize the domain gap; $r_{\mathrm{eval}}{=}2.0$\,m is the ICP correspondence cutoff for RMSE evaluation.
Point-to-plane ICP requires reliable surface normals, which are poorly estimated at low aerial densities (2--8\,pts/m$^2$; Sec.~\ref{sec:dataset}); we therefore use point-to-point estimation throughout.
All geometric methods run on a single CPU core (Intel Xeon); learned baselines use GPU inference (RTX 3090) with CPU-only ICP refinement (Open3D 0.18.0~\cite{open3d}).

\smallskip\noindent\emph{Implementation details.}
All ICP stages use native resolution (no voxel downsampling);
ground scans contain ${\approx}$10--30K points (VLP-16 ${\approx}$10K,
OS1-64 ${\approx}$30K); aerial crops contain ${\approx}$5--50K points
within the 50\,m radius. ICP uses Open3D point-to-point estimation
with the stage-specific max correspondence distances above, 50 iterations
per stage; candidate scoring uses $r_{\mathrm{eval}}{=}2.0$\,m.
BEV rasterization: $200{\times}200$ at 0.5\,m/pixel;
FM: $K_\theta{=}3$ yaw, $K_t{=}5$ translations (${\leq}$15 total).
FPFH-RANSAC: voxel downsample 0.5\,m, FPFH radius 1.0\,m, max correspondence 2.0\,m, 4M iterations (Open3D defaults otherwise), refined by CTF.

\smallskip\noindent\fbox{\parbox{\dimexpr\columnwidth-2\fboxsep-2\fboxrule}{%
\textbf{Primary benchmark:} 9{,}012 scans, 6 sites, Protocol~B ($\pm$5\,m, $\pm$15$^\circ$).
Metrics: S@1.0\,m (coarse RMSE convergence), S@0.75\,m (primary sub-meter target), S@0.5\,m (floor; limited by low aerial density).
All RGSR results use this setting (Table~\ref{tab:rgsr}); all methods share identical per-scan jitter draws (seed 42).}}\smallskip

\emph{Baselines} (7 methods).
(1)~Standard CTF,
(2)~Two-Stage CTF ($p{=}30$),
(3)~GICP~\cite{gicp}+CTF,
(4)~Trimmed ICP~\cite{tricp}+CTF,
(5)~FGR~\cite{fgr}+ICP,
(6)~GeoTransformer~\cite{geotransformer}+CTF (trained on CSLiDAR),
(7)~BUFFER-X~\cite{bufferx} zero-shot (3DMatch~\cite{threedmatch}).
Methods 1--2 use the full geometric benchmarks. GeoTransformer is full-scale in Table~\ref{tab:rgsr} and canonical-subset in Table~\ref{tab:cascade}; methods 3--5 and BUFFER-X use trajectory-balanced subsets (200\,scans/traj., Protocol~B), yielding GICP+CTF 58\%, Trimmed~ICP+CTF 50\%, and FGR+ICP 74.6\% S@0.75\,m vs.\ 71\% CTF (absolute S@$\tau$ lower due to uniform trajectory weighting); robust estimation degrades at low aerial density.
TEASER++~\cite{teaser} and Super4PCS~\cite{super4pcs} depend on reliable cross-source correspondences.
Our geometric methods use no training; we tuned $\tau_g$ and the percentile set $\{15,30,45,60\}$ on CUA and Georgetown development runs under Protocol~B, then froze all parameters (CTF schedule, $r_{\mathrm{eval}}$, Phase~2 gates fixed a priori; GMU, GWU, and UMD not used for selection).
GeoTransformer was trained from scratch on CSLiDAR (official code, voxel 0.3\,m, $\pm$2\,m jitter); BUFFER-X uses the released 3DMatch checkpoint zero-shot; we were unable to obtain CrossPCR~\cite{crosspcr} or SPEAL~\cite{speal} implementations.

\subsection{Results: Height Stratification}

Table~\ref{tab:cascade} shows the progression from standard CTF through our height-stratified methods under Protocol~A.
Standard CTF achieves 97.1\% S@1.0\,m but only 78.8\% S@0.75\,m, indicating many scans converge to nearby but imprecise solutions.
Two-Stage CTF ($p{=}30$) improves S@0.75\,m to 80.5\% ($+$1.7\,pp) by bootstrapping from ground-plane correspondences.
The confidence-gated cascade reaches \textbf{83.1\%} S@0.75\,m and \textbf{99.0\%} S@1.0\,m on all 12,683 scans by thresholding escalation through CTF $\to$ Two-Stage $\to$ RANSAC+CTF using RMSE thresholds.

Per-site gains under Protocol~B (Table~\ref{tab:rgsr}) confirm the cascade's advantage concentrates on lower-overlap sites.

\begin{table}[t]
\centering
\caption{\textbf{Method progression under Protocol~A} (reference init). Top block: geometric methods on all \textbf{12{,}683} scans. Bottom block: learned baselines on a stratified canonical subset (see footnotes); all methods use the same $e(T)$ metric ($r_{\mathrm{eval}}{=}2.0\,\mathrm{m}$). Height stratification improves sub-meter success with modest runtime overhead.}
\label{tab:cascade}
\small
\setlength{\tabcolsep}{2pt}
\begin{tabular}{@{}l rr r@{}}
\toprule
Method & S@0.75\,\textnormal{m} (\%) & S@1.0\,\textnormal{m} (\%) & Time$^*$ \\
\midrule
Standard CTF & 78.8 & 97.1 & 1.0$\times$ \\
Two-Stage CTF ($p{=}30$) & 80.5 & 97.4 & 1.1$\times$ \\
Cascade ($\tau_g{=}0.75$\,m) & \textbf{83.1} & \textbf{99.0} & 1.2$\times$ \\
\midrule
GeoTransformer$^\dagger$ + CTF & 69.4 & 93.1 & ${\sim}$9$\times$ \\
BUFFER-X$^\ddagger$ (zero-shot) & 32.8 & 71.7 & ${\sim}$2$\times$ \\
\bottomrule
\multicolumn{4}{@{}p{\columnwidth}@{}}{\scriptsize $^\dagger$Trained on CSLiDAR (all campuses; in-domain supervised baseline); 1{,}600 canonical scans$^{\S}$ (full-scale: 76.3\% S@0.75\,m on 9{,}012 scans; Table~\ref{tab:rgsr}).}\\
\multicolumn{4}{@{}p{\columnwidth}@{}}{\scriptsize $^\ddagger$Pre-trained (3DMatch) zero-shot; 1{,}200 canonical scans$^{\S}$.}\\
\multicolumn{4}{@{}p{\columnwidth}@{}}{\scriptsize $^{\S}$Canonical: 200\,scans/traj.\ (uniform stride) from the 8 trajectories retained for \mbox{Protocol~B} (6 for~$^\ddagger$); \mbox{Protocol~A} init; seed~42.}\\
\multicolumn{4}{@{}p{\columnwidth}@{}}{\scriptsize $^*$Relative to Standard CTF (CPU); $^\dagger\!$,$^\ddagger$ use GPU inference.}
\end{tabular}
\end{table}

\subsection{Results: Learned Baselines}

GeoTransformer trained on CSLiDAR ($T_{\mathrm{init}}$-independent; CTF seeded by its estimated transform) achieves 76.3\% S@0.75\,m on all 9{,}012 Protocol~B scans (Table~\ref{tab:rgsr}), comparable to standalone CTF (76.7\%) but 9.7\,pp below RGSR; on canonical subsets it reaches 69.4\% (Table~\ref{tab:cascade}).
BUFFER-X (zero-shot) reaches only 32.8\% S@0.75\,m; both results confirm that coverage asymmetry degrades learned correspondences regardless of training domain.

\subsection{Results: Primary Benchmark}

Table~\ref{tab:rgsr} compares the geometric portfolio and GeoTransformer (full-scale, 9{,}012 scans) under Protocol~B.
GeoTransformer matches standalone CTF in aggregate (76.3\% vs.\ 76.7\%) but collapses on low-overlap sites: 11.4\% on UMD-IdeaFactory (Cov@1\,m${=}$20\%) and 55.6\% on UMD-Iribe (46\%), vs.\ ${\approx}$92--95\% on higher-coverage GMU/CUA.
The confidence-gated cascade reaches 83.7\%; \textbf{RGSR} extends to \textbf{86.0\%} S@0.75\,m and \textbf{99.8\%} S@1.0\,m, outperforming GeoTransformer by 9.7\,pp overall and by 16--34\,pp on hard sites (0 RMSE-regressed; pose validity supported by survey control and trajectory consistency, Sec.~\ref{sec:analysis}).
On sites not used for tuning (UMD, GMU, GWU; 5{,}616 scans), RGSR achieves 83.9\% vs.\ 73.4\% for GeoTransformer+CTF ($+$10.5\,pp).
Remaining failures concentrate in lower-overlap sites (UMD, Georgetown; Cov@1\,m 20--61\%), which comprise 33\% of scans but 87\% of failures at S@0.75\,m; sub-canopy and facade geometry limits convergence quality regardless of initialization diversity.
An exploratory +FM extension raises RMSE-based success to 88.8\% but increases TRE on survey-control scans (Sec.~\ref{sec:analysis}); we report RGSR as the validated primary result.

\begin{table}[t]
\centering
\caption{\textbf{Portfolio results} (Protocol~B, $\pm 5\,\mathrm{m}/\pm 15^{\circ}$, $\tau_g{=}0.75\,\mathrm{m}$). S@$\tau$: fraction with inlier RMSE${<}\tau$. GeoTr.: GeoTransformer (CSLiDAR-trained) + CTF (Sec.~\ref{sec:experiments}); Casc.: confidence-gated cascade; RGSR: Phases~1--3 (Sec.~\ref{sec:rgsr}). Accept-if-better: 0 RGSR regressions vs.\ cascade. Baselines: CTF 76.7\%; Two-Stage 77.9\%. Gains are largest on lower-coverage sites (UMD, Georgetown).}
\label{tab:rgsr}
\small
\setlength{\tabcolsep}{3pt}
\begin{tabular}{l r rrr}
\toprule
 & & \multicolumn{3}{c}{S@0.75\,\textnormal{m} (\%)} \\
\cmidrule(lr){3-5}
Site (Cov@1\,m) & $n$ & GeoTr.$^\dagger$ & Casc. & \textbf{RGSR} \\
\midrule
Georgetown (61\%) & 1,069 & 57.2 & 70.7 & \textbf{73.3} \\
CUA (68\%) & 2,327 & 92.1 & 95.9 & \textbf{97.1} \\
UMD-IdeaF.\ (20\%) & 1,031 & 11.4 & 35.5 & \textbf{45.0} \\
UMD-Iribe (46\%) & 863 & 55.6 & 68.1 & \textbf{71.6} \\
\cmidrule(lr){1-5}
GMU (65\%) & 3,190 & 95.2 & 96.4 & \textbf{97.3} \\
GWU (66\%) & 532 & 91.4 & 98.5 & \textbf{98.9} \\
\midrule
\textbf{All 6 sites} & \textbf{9,012} & 76.3 & 83.7 & \textbf{86.0} \\
\midrule
\multicolumn{5}{@{}p{\columnwidth}@{}}{\scriptsize\itshape Other thresholds (all 9{,}012):} \\
\multicolumn{2}{l}{S@0.5\,\textnormal{m}} & \textbf{3.5} & 2.9 & 3.0 \\
\multicolumn{2}{l}{S@1.0\,\textnormal{m}} & 95.0 & 99.5 & \textbf{99.8} \\
\bottomrule
\multicolumn{5}{@{}p{\columnwidth}@{}}{\scriptsize $^\dagger$GeoTransformer predicts correspondences from the scan pair; the resulting estimated transform seeds CTF refinement (independent of $T_{\mathrm{init}}$).} \\
\multicolumn{5}{@{}p{\columnwidth}@{}}{\scriptsize \phantom{$^\dagger$}BUFFER-X (zero-shot): 32.8\% S@0.75\,m (1{,}200 scans).} \\
\multicolumn{5}{@{}p{\columnwidth}@{}}{\scriptsize \phantom{$^\dagger$}+FM (abl.): 88.8\% S@0.75\,m but increases TRE (Sec.~\ref{sec:analysis}).}
\end{tabular}
\end{table}

\subsection{Analysis}
\label{sec:analysis}

\begin{figure}[t]
  \centering
  \includegraphics[width=\columnwidth]{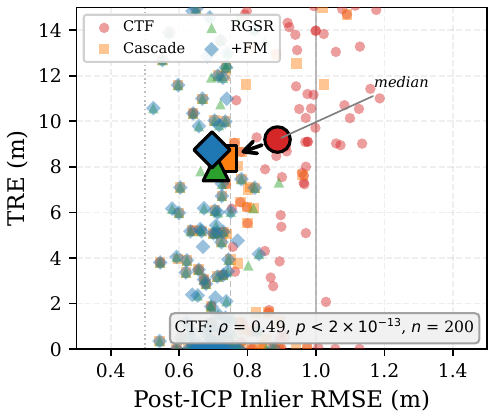}
  \caption{%
    \textbf{RMSE vs.\ translation recovery error (TRE)} on 200 UMD scans (Protocol~B) across four pipeline stages.
    Per-scan TRE is the median $\|T_{\mathrm{est}}\,p_k{-}p_k^{\mathrm{ref}}\|$ over survey markers $p_k$; each point is one (scan, stage) pair.
    Vertical lines mark S@$\tau$ thresholds (0.5, 0.75, 1.0\,m);
    large markers with arrows trace stage-wise medians.
    Spearman $\rho{=}0.49$, $p$-value${<}2{\times}10^{-13}$ (CTF stage, $n{=}200$; reported before hypothesis selection to avoid selection-bias inflation).
    Through RGSR, successive stages compress RMSE while reducing TRE (stage-wise median TRE: 9.2$\to$8.4$\to$7.95\,m).
    The +FM stage (diamonds) further lowers RMSE but \emph{increases} median TRE to 8.75\,m, illustrating the RMSE--TRE disconnect under extreme partial overlap and motivating independent pose validation.
  }
  \label{fig:rmse_tre}
\end{figure}

\textbf{Diagnostic findings and non-circular validation.}
Post-ICP inlier RMSE $e(T)$ is used \emph{only} for selection/escalation; S@$\tau$ is non-decreasing by construction (accept-if-better).
However, RMSE monotonicity does not guarantee monotone pose improvement: the inlier set $\mathcal{I}(T)$ changes with each transform, so RMSE can decrease by convergence to a different local minimum rather than correcting pose.
We validate with two independent measurements:
\textbf{(i)~Survey control} ($N{=}200$ UMD scans).
To validate that lower $e(T)$ corresponds to lower pose error, we evaluate against total-station survey markers via translation recovery error $\mathrm{TRE}$: the per-scan median $\|T_{\mathrm{est}}\,p_k{-}p_k^{\mathrm{ref}}\|$ over markers $p_k$ (independent of the aerial map).
At CTF stage, RMSE and TRE are moderately correlated (Spearman $\rho{=}0.49$, $p$-value${<}2{\times}10^{-13}$; Fig.~\ref{fig:rmse_tre}); scans with RMSE${<}0.75$\,m ($n{=}57$) have 0.10\,m median TRE.
On these 200 scans, S@0.75\,m increases stage-wise (28.5\%$\to$61.5\%$\to$81.0\% for CTF$\to$Cascade$\to$RGSR) while median TRE over all scans (including failures) decreases at each stage (9.2$\to$8.4$\to$7.95\,m), confirming that the RGSR hypothesis set improves pose, not just RMSE.
The exploratory +FM extension further raises S@0.75\,m on this survey subset to 90.5\% but \emph{increases} median TRE to 8.75\,m, demonstrating that additional proposals can produce spurious low-RMSE local minima under extreme partial overlap.
\textbf{(ii)~Trajectory consistency:} Local Motion Consistency Error (LMCE)---the per-scan norm $\|t(T_i^{-1}T_{i+1})-t(O_i^{-1}O_{i+1})\|$, where $T_i$ is the refined pose and $O_i$ the SLAM odometry pose---median drops from 5.0\,m (jittered init) to 0.49\,m after RGSR on 3 trajectories, confirming lower RMSE yields odometry-consistent poses.

\emph{Compute.}
Single Xeon core: CTF~1.0\,s; RGSR~3.0\,s mean; under reference initialization (Protocol~A) the selected output stage is CTF for 95.7\% of scans (2.5\% are selected at Two-Stage, 1.8\% at RANSAC).
Cascade S@0.75\,m varies by ${<}$0.1\,pp across $\tau_g \in [0.5, 0.75]$\,m (Protocol~A, 12{,}683 scans).

\textbf{Reverse-direction ablation.}
Disabling reverse-direction hypotheses (Phase~1) on all 9{,}012 Protocol~B scans reduces S@0.75\,m by 3.4\,pp; the gap concentrates on the hardest sites (IdeaFactory $-$14.3\,pp, Iribe $-$7.5\,pp, Georgetown $-$4.6\,pp), confirming reverse-direction ICP exploits high aerial-to-ground coverage to reach local minima inaccessible to forward ICP alone.

\emph{Jitter sensitivity.}
On a stratified subset (1{,}400 scans, 200/traj.${\times}$7, excluding near-ceiling GWU), RGSR S@0.75\,m varies by ${<}$0.2\,pp across a 5$\times$ jitter range ($\pm$2\,m/$\pm$5$^\circ$ to $\pm$10\,m/$\pm$30$^\circ$); the cascade baseline is similarly stable.

\emph{Metric saturation.}
RGSR S@0.5\,m is only 3.0\% (Table~\ref{tab:rgsr}), but even the reference alignment achieves only 0.1\% S@0.5\,m on the full 9{,}012 scans (cross-source NN residuals have a nonzero floor at $T_{\mathrm{ref}}$ due to modality mismatch), suggesting the 0.5\,m floor reflects low aerial density (2--8\,pts/m$^2$; Sec.~\ref{sec:dataset}) and reference alignment accuracy, not method failure.

\textbf{RMSE--TRE disconnect.}
Pseudo-pose-error validation across all 9{,}012 Protocol~B scans confirms this disconnect scales with coverage: on high-coverage sites (Cov@1\,m${\geq}$65\%), RGSR median pose deviation stays within 0.04\,m of CTF, while on low-coverage sites (20--61\%) it increases by 0.5--3.1\,m despite S@0.75\,m gains of 10--34\,pp.
Independent pose validation is essential for any RMSE-based selection pipeline applied to cross-source registration.

\section{DISCUSSION AND LIMITATIONS}
\label{sec:discussion}

\textbf{Discussion.}
Under extreme aerial--ground coverage asymmetry, \emph{local-minimum selection} is a key challenge: RGSR's height-stratified and reverse-direction hypotheses outperform learned baselines on Paired-CSLiDAR, though the RMSE--pose relationship weakens at lower coverage (Sec.~\ref{sec:analysis}).
In the refinement setting studied here, RGSR demonstrates pose refinement against publicly available airborne LiDAR maps (e.g., USGS 3DEP~\cite{usgs3dep}) at ${\approx}$3\,s/scan on a single CPU core; Paired-CSLiDAR provides per-scan coverage statistics and reference alignments to facilitate method comparison by overlap level.

\textbf{Limitations.}
RGSR averages ${\approx}$3\,s/scan on a single CPU core; Phase~1 hypotheses are independent and parallelizable.
The benchmark assumes a correct 50\,m crop; we do not evaluate wrong-crop rejection, and on incorrect crops both high-RMSE failures and spuriously low-RMSE alignments (from structural similarity) may occur, motivating an explicit crop-verification stage.
Protocol~B perturbs only $(x, y, \text{yaw})$ (gravity-aligned); robustness to attitude/elevation errors or non-planar terrain is not evaluated.
Remaining failures tend to occur in observability-limited settings (dense canopy, open areas with weak lateral structure), where multi-scan or temporal context may help.
Finally, most scans use \emph{reference} (not ground-truth) alignments; independent survey control covers 200 UMD scans, and full-scale pseudo-pose-error analysis shows the RMSE--pose disconnect scales with coverage (Sec.~\ref{sec:analysis}); all sites are US mid-Atlantic and temporal change is not modeled.

\section{CONCLUSION}

We introduced Paired-CSLiDAR: 12{,}683 ground--aerial pairs across 6 evaluation sites with per-scan reference SE(3) alignments for single-scan metric pose refinement.
\mbox{Using} shared ground-plane structure, our geometry-only height-stratified, non-regressive cascade reaches 83.1\% S@0.75\,m under Protocol~A; RGSR adds multi-percentile and reverse-direction hypotheses plus residual-guided refinement to achieve \textbf{86.0\%} S@0.75\,m and~\textbf{99.8\%} S@1.0\,m on 9{,}012 Protocol~B scans, with RMSE-based selection supported by survey control and trajectory consistency.
An exploratory Fourier--Mellin BEV extension (+FM) shows RMSE can improve while TRE worsens under extreme partial overlap, motivating independent pose validation when adding new proposal mechanisms.
The dataset (CC~BY~4.0) and code (MIT) are being prepared for public release.


\textbf{AI Disclosure:}
ChatGPT was used for language editing (grammar/clarity) on the Abstract and Sections~I--VII (text only). All technical content, experimental design, implementation, and result interpretation were produced and verified by the authors.

\begingroup\let\footnotesize\scriptsize

\endgroup

\end{document}